%% file: main.tex
\documentclass[10pt,twocolumn,letterpaper]{article}

\usepackage[pagenumbers]{iccv} 


\definecolor{iccvblue}{rgb}{0.21,0.49,0.74}
\usepackage[pagebackref,breaklinks,colorlinks,allcolors=iccvblue]{hyperref}

\usepackage{url}

\usepackage{algorithm}
\usepackage{algorithmic}
\usepackage{booktabs}
\usepackage{times}
\usepackage{epsfig}
\usepackage{amsmath}
\usepackage{amssymb}
\usepackage{caption}
\usepackage{multirow}
\usepackage[flushleft]{threeparttable}
\usepackage{pifont}

\usepackage{color}
\usepackage{array}
\usepackage{makecell}

\usepackage{comment}

\newlength\savewidth
\usepackage[utf8]{inputenc} 
\usepackage[T1]{fontenc}    
\usepackage{hyperref}       
\usepackage{amsfonts}       
\usepackage{nicefrac}       
\usepackage{microtype}      
\usepackage{xcolor}         

\usepackage{tabularray}
\usepackage{colortbl}
\usepackage{arydshln}

\definecolor{mygray}{gray}{.9}
\definecolor{mygreen}{rgb}{0, 0.6, 0}

\def\paperID{13263} 
\def\confName{ICCV}
\def\confYear{2025}

\title{Scaling Open-Vocabulary Action Detection}

\author{Zhen Hao Sia \quad Yogesh Singh Rawat \\
{\tt\small \{zhenhao.sia, yogesh\}@ucf.edu} \\
University of Central Florida \\
}

\begin{document}
\maketitle
\input{sec/0_abstract}
\input{sec/1_intro}
\input{sec/2_related}
\input{sec/3_method}
\input{sec/4_exp}
\input{sec/5_conclusion}

{
    \small
    \bibliographystyle{ieeenat_fullname}
    \bibliography{main}
}

\end{document}

%% file: sec/0_abstract.tex
\begin{abstract}
In this work, we focus on scaling open-vocabulary action detection. Existing approaches for action detection are predominantly limited to closed-set scenarios and rely on complex, parameter-heavy architectures. Extending these models to the open-vocabulary setting poses two key challenges: (1) the lack of large-scale datasets with many action classes for robust training, and (2) parameter-heavy adaptations to a pretrained vision-language contrastive model to convert it for detection, risking overfitting the additional non-pretrained parameters to base action classes. Firstly, we introduce an encoder-only multimodal model for video action detection, reducing the reliance on parameter-heavy additions for video action detection. Secondly, we introduce a simple weakly supervised training strategy to exploit an existing closed-set action detection dataset for pretraining. Finally, we depart from the ill-posed base-to-novel benchmark used by prior works in open-vocabulary action detection and devise a new benchmark to evaluate on existing closed-set action detection datasets without ever using them for training, showing novel results to serve as baselines for future work. Our code is available at \url{https://siatheindochinese.github.io/sia_act_page/}.
\end{abstract}

\begin{figure} 
    \centering
    \includegraphics[scale=0.33]{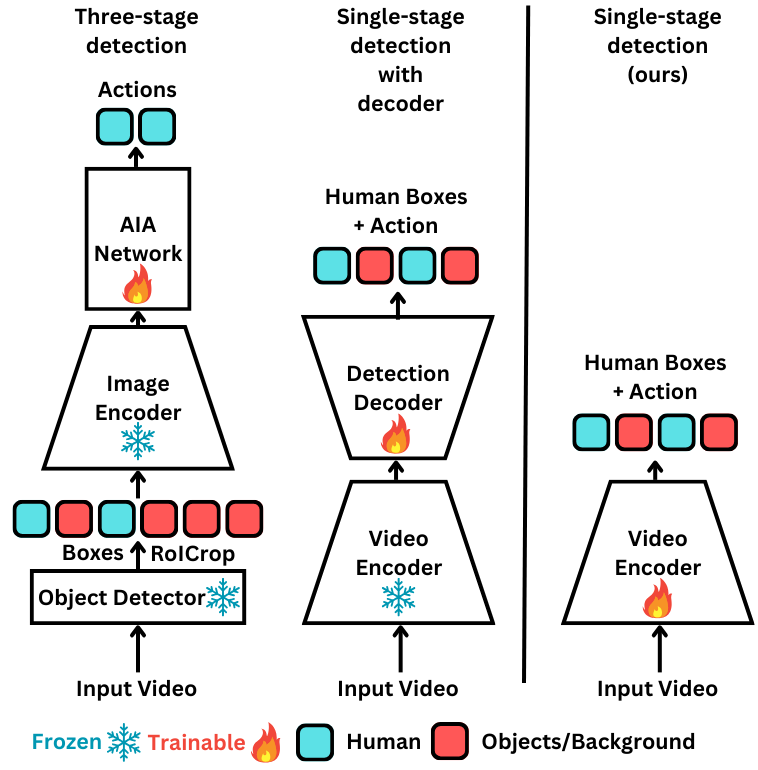}
    \caption{\textbf{\textit{Overview of existing approaches}} to convert pretrained vision/video-language models for action detection (left, middle) vs ours (right).
    }
    \label{fig:teaser}
    \vspace{-10pt}
\end{figure}

%% file: sec/1_intro.tex
\section{Introduction}
\label{sec:intro}

Spatiotemporal action detection has traditionally focused on the closed-set scenario, where models are trained on fully-supervised, predefined action classes and can only recognize actions encountered during training. While effective within controlled environments, this approach is restrictive for real-world applications, where human actions and interactions are inherently diverse and constantly evolving. From public surveillance and autonomous systems to sports analytics and assistive technologies, the range of possible actions is vast and unpredictable, often extending far beyond the limited set captured in any single dataset. Open-vocabulary action detection addresses this limitation by allowing models to detect and identify novel action classes unseen during training, providing a more scalable and adaptable solution for real-world scenarios.

Developing open-vocabulary models for action detection is challenging due to two main factors: (1) while vision-language modeling offers a promising approach to extend closed-set models \cite{evad, star, tuber, stmixer, hit, bmvit} for detecting novel actions, naively applying this to action detection remains challenging due to the scarcity of annotated datasets covering a large number of actions, as opposed to open-vocabulary object detection where large-scale datasets with sufficient number of object classes already exist \cite{lvis}, and (2) while existing work \cite{iclip, openmixer} tries to alleviate this issue by adapting a pretrained, frozen vision/video-language contrastive model inside a detection architecture, having a significant increase in the number of non-pretrained modules/parameters leads to the risk that the additional parameters overfit to the base action classes, as well as incurring additional computation.

In this work, we address these two challenges by (1) proposing a weakly-supervised approach to significantly inflate the number of action classes seen during pretraining to deal with the scarcity of action classes, and (2) introducing a single-stage, encoder-only, open-vocabulary action detection model that does not rely on parameter-heavy additions such as a decoder or an external human detector to avoid overfitting these new parameters to base action classes (Figure \ref{fig:teaser}).

The abundance of action classes from our training scheme allows us to avoid freezing pretrained vision/video-language models, enabling our model to adapt more effectively to novel actions without sacrificing generalization. Additionally, by avoiding parameter-heavy modules attached to a frozen vision/video encoder, our approach remains lightweight, reducing computational overhead and making it more practical for real-world deployment, as well as allowing it to be trained end-to-end.

We perform extensive experiments on six different existing action detection datasets: AVA \cite{ava} , AVA-Kinetics (AVA-K) \cite{avak},  UCF-101-24 \cite{ucf101}, JHMDB51-21 \cite{jhmdb}, Multisports \cite{multisports}, and UCF-MAMA \cite{ucfmama}, demonstrating the open-vocabulary capability of our model.

In summary, we have the following contributions:
\begin{itemize}
    \item We introduce SiA, a \textit{si}mple \textit{a}rchitecture for open-vocabulary action detection which is multimodal and lightweight.
    \item We introduce a weakly-supervised training scheme to exploit the largest existing closed-set action detection dataset by inflating the number of action classes seen during training from 80 to more than 700, a significant departure from prior work that has only seen less than 20 actions during pretraining.
    \item We depart from the ill-posed base-to-novel benchmark used by prior works in zero-shot/open-vocabulary action detection and set a new benchmark for the task of open-vocabulary action detection by showing open-vocabulary results on UCF-101-24, JHMDB, MultiSports and UCF-MAMA without training our model on these datasets, setting a novel baseline.
\end{itemize}

%% file: sec/2_related.tex
\section{Related Work}

\noindent \textbf{Spatio-Temporal Action Detection} fall under two categories: frame-level \cite{stadsurvey} and clip-level \cite{stadsurvey}. Clip-level action detectors output spatiotemporal tubelets with their corresponding action in a given video clip \cite{stadsurvey}, whereas frame-level action detectors output human boxes and their actions only for a given keyframe in a video clip \cite{stadsurvey}, relying on postprocessing methods to build spatiotemporal tubelets. Clip-level action detection is usually more computationally expensive compared to frame-level action detection. We focus on frame-level task in this work.

\noindent \textbf{Adapting Recognition/Classification Transformer Backbones for Detection} involve either: (1) directly regressing [PATCH] tokens \cite{owlvit, owlv2, bmvit}, (2) adding an extra sequence of trainable [DET] detection tokens to the input and regressing those tokens at the output \cite{yolos}, (3) using them as a backbone to produce feature maps, following a two-stage/FasterRCNN detection scheme \cite{slowfast, videomae, videomaev2}, and (4) using them as the encoder in an encoder-decoder architecture similar to DETR or AdaMixer \cite{tuber, stmixer, star}. As of current, (2) has not yet been explored for video transformers for action detection, nor has it been explored for multimodality; we explore this scheme for videos as well as its utility in the open-vocabulary setting.

\noindent \textbf{Open-Vocabulary Learning} has dominated tasks in the image domain, such as image classification \cite{clip}, object detection \cite{owlvit, owlv2}, and image segmentation \cite{sam}. In the video domain, most open-vocabulary works revolve around video classification/retrieval \cite{internvid, xclip, vificlip, clipvip, froster, ezclip, actionclip} and specifically for human actions, temporal action detection \cite{ZSTAL, TALbaseline, itcetad, mvavta} which only localizes the start and end times of an action, not the spatial location of people and their individual actions. In the more spatially fine-grained task of action detection in videos, existing works are predominantly closed-set.

Open-vocabulary action detection remains an ill-posed task primarily due to the lack of large-scale action detection datasets. Commonly used action detection datasets, such as UCF-101-24 \cite{ucf101}, JHMDB \cite{jhmdb}, AVA \cite{ava}, and AVA-Kinetics \cite{avak} have between 21-80 action classes. Prior methods to deal with the lack of large scale data revolve around adapting existing vision-language contrastive models which are already pretrained on large image-text/video-text datasets for action detection and using an ill-posed base-to-novel scheme to split the datasets into base actions for training and novel actions for evaluation.

iCLIP \cite{iclip} is one of the first work to extend action detection into the vision-language domain by adapting frozen CLIP \cite{clip} image and text encoders within an Asynchronous Interaction Aggregation (AIA) network \cite{aia}. The model employs a complicated pipeline: frames from an input video clip are processed by a pretrained closed-set object detector, then object proposals are cropped and encoded by the frozen CLIP image encoder before passing through the AIA network. This multi-stage design introduces inefficiencies and a key bottleneck: the closed-set object detector. Since this detector is limited to objects it was trained on, it may fail to capture novel or unlabeled objects, leading to potential information loss. OpenMixer \cite{openmixer} extends the closed-set encoder-decoder action detection model STMixer \cite{stmixer} to the open-vocabulary setting by adding a frozen video backbone and text encoder from a video-language contrastive model.

These existing approaches are pretrained on small-scale action detection datasets, limiting their exposure to a maximum of only 18 action classes during training, which risks overfitting the model to these actions.

In contrast, (1) our training method allows more than 700 action classes to be seen during training, leading to (2) not having to rely on freezing the pretrained vision and language encoders to preserve learned semantics, and (3) our model is encoder-only; we do not rely on adding an external human detector or a parameter-heavy decoder.

\noindent \textbf{Scaling Open-Vocabulary Detection with Weak Supervision} in the task of object detection in images involve extending detection capabilities to more classes without exhaustive manual labeling. DETIC \cite{detic} introduces a method to use image classification datasets that have no annotated bounding boxes by implementing several heuristics to generate a pseudobox for each image. OWLv2 \cite{owlv2}, 3Ways \cite{3ways} and RegionCLIP \cite{regionclip} rely on self-training by collecting pseudoboxes from their own detections on large-scale image-text datasets and further training their model on these boxes.

In contrast, our method does not increase the number of videos, nor do we use pseudoboxes. To our knowledge, there is no existing work focused on weakly-supervised scaling for open-vocabulary action detection.

\begin{figure*} 
    \centering
    \includegraphics[scale=0.344]{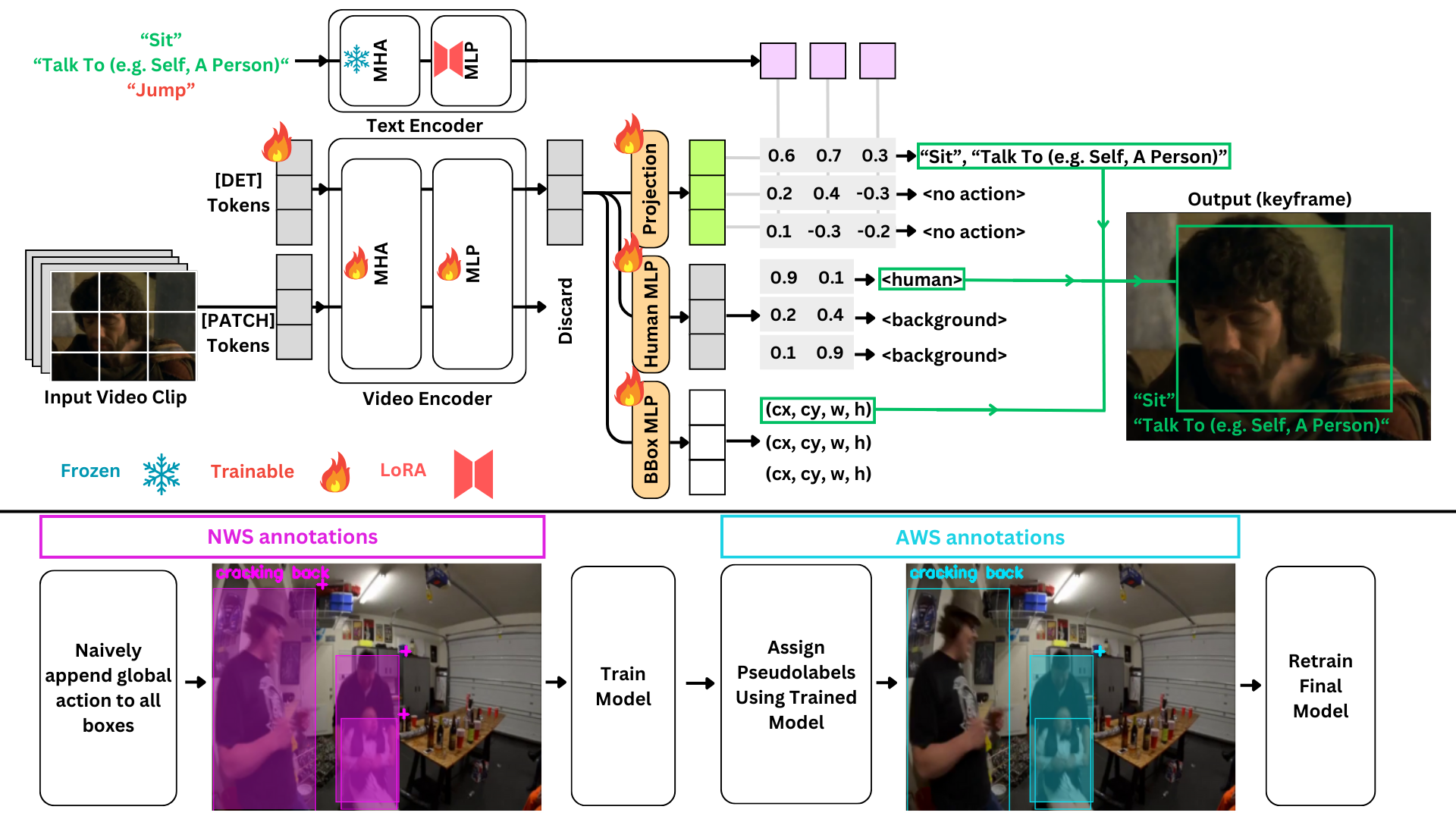}
    \caption{\textbf{\textit{Overview of SiA architecture (top) and our weak-supervision scheme (bottom):}} Our architecture consists of a contrastive-pretrained video and text encoder. We add [DET] tokens to the input sequence and regress them at the output to convert the model for action detection. We train it to specifically detect humans and their actions only in the keyframe, which we set to the middle frame in any given input video clip. Our weak-supervision scheme naively appends the global action of Kinetics-700 videos to their AVA annotations; refined annotations are obtained by using our model trained on these naive annotations to assign the global action to the right boxes. In the shown example, the highlighted boxes denotes that the global action has been assigned to the human box; the Kinetics-700 action `cracking back' is properly assigned to the only person cracking the back of another person.
    }
    \label{fig:contrib}
    \vspace{-10pt}
\end{figure*}

%% file: sec/3_method.tex
\section{Methodology}
\label{sec:method}

\noindent \textbf{Problem Formulation} Given a video $V = (v_{1}, v_{2}, ...v_{\textit{L}})$ with $L$ frames, the task of frame-level action detection is to train a model to output a set of human bounding boxes on the keyframe $v_{K}$ and classify the actions associated with them.
We adopt the open-vocabulary definition from the well-studied object detection problem in images \cite{owlvit}, where the model is trained on a set of base action classes and is expected to generalize to both base and novel action classes during testing.
In the following sections, we introduce our model and our weakly supervised training scheme centered on AVA-Kinetics to expand the number of action classes. An overview of our approach is shown in Figure \ref{fig:contrib}.

\subsection{SiA Architecture}
\label{sec:archi}

Our model consists of a video encoder and text encoder which has been initially contrastively pretrained on a large scale video-text dataset for open-vocabulary video classification/retrieval.

\noindent \textbf{Video Encoder:} In the interest of avoiding overfitting newly added parameters to base action classes, we seek to avoid designing and attaching additional parameter-heavy modules to the pretrained video encoder. We attempt two schemes on the video encoder for action detection as following:

\begin{enumerate}
    \item Temporally average pool [PATCH] tokens at the output and regress them for action detection, which has already been explored by BMViT \cite{bmvit}.
    \item Remove the [CLS] token and add 100 [DET] tokens to the input sequence and regress them at the output for action detection. To our knowledge, this scheme has not yet been attempted for video action detection.
\end{enumerate}

In our ablations, we find that the [DET] token scheme is a better alternative, and choose it for our final model.

Following transformer-based detection architectures \cite{detr, yolos, bmvit, star, tuber, vidt, stmixer}, we use two MLPs to regress the output tokens to obtain bounding boxes and actor scores, and a projection layer projects these tokens into vision embeddings; for each output token, the modified video encoder outputs a triplet of bounding box coordinates, actor probability and a vision embedding, $(\textbf{b}, \textbf{p}_{act}, \textbf{e}_{v})$. Tokens with $\textbf{p}_{act}$ more than 0.5 are considered to have a actor in them, whereas tokens with $\textbf{p}_{act}$ less than 0.5 are considered background tokens and filtered out.

\noindent \textbf{Text Encoder:} We utilize LoRA \cite{lora} for the MLPs in each transformer block of the text encoder, keeping the original weights frozen and finetuning the LoRA modules to better align the output text embeddings $embed_{t}$ for region-specific actions. We show the importance of LoRA-finetuning the text encoder in our ablations.

\noindent \textbf{Detecting Actions:} For any given input clip, we designate the middle frame as the keyframe and specifically train our model to detect humans and their actions within that frame.
For each output token with a positive human detection at the output of the video encoder, we determine the actions of the detected individual by calculating the cosine similarity $S$ between $\textbf{e}_{v}$, and the encoded text embedding, $\textbf{e}_{t}$, of the target action, where $S = \frac{\textbf{e}_{v} \cdot \textbf{e}_{t}}{|\textbf{e}_{v}||\textbf{e}_{t}|}$. $S$ closer to 1 indicates a higher likelihood that the detected person is performing the specified target action, and vice versa for values close to -1.

The text encoder is able to encode any action as a textual input, expanding detection capabilities to actions unseen during training, provided that a sufficient number of human actions are used during training.
In contrast to previous models (iCLIP and OpenMixer), our model is single-stage and end-to-end trainable, avoiding the need for additional parameter-heavy modules attached to the vision encoder or external detectors to generate human proposals.

\subsection{Scaling Action Classes Using Weak Supervision}

\begin{table}[t]
\centering
\caption{\textbf{\textit{Number of action classes}} in existing closed-set action detection datasets versus after both of our weakly-supervised training recipes: Naive Weak Supervision (NWS) and Assignment-based Weak Supervision (AWS). (UCF-MAMA-H: only human actions)}
\renewcommand{\arraystretch}{1.06}
\scalebox{0.9}{
    \begin{tabular}{r|c c}
        \rowcolor{mygray} 
        \specialrule{1.5pt}{0pt}{0pt}
        Dataset & \# Action Classes & Multi-label Actions \\
        JHMDB & 21 & $\times$ \\
        UCF-101-24 & 24 & $\times$ \\     
        UCF-MAMA & 35 & $\times$ \\
        UCF-MAMA-H & 27 & $\times$ \\
        MultiSports & 66 & $\times$ \\
        AVA & 80 & \checkmark \\
        \hline
        AVA-Kinetics & 80 & \checkmark \\
        +NWS (ours) & 700+ & \checkmark \\
        +AWS (ours) & 700+ & \checkmark \\
        \specialrule{1.5pt}{0pt}{0pt}
    \end{tabular}}
\label{tab:numact}
\vspace{-10pt}
\end{table}

To address the limited number of action classes in existing action detection datasets as shown in Table \ref{tab:numact}, we aim to significantly increase the number of action classes by exploiting and unlocking the full potential of AVA-Kinetics \cite{avak} with two weakly-supervised approaches outlined below:

\noindent
\textbf{Naive Weak Supervision (NWS):} AVA-Kinetics is a dataset that combines the original AVA dataset with a subset of Kinetics-700 \cite{k700} videos, annotated with the 80 action classes of AVA. Since the human boxes in these Kinetics-700 videos are already annotated following the AVA format with AVA classes, we introduce a weakly-supervised approach to expand the number of action classes from 80 to over 700. This is achieved by appending the global action label of each Kinetics-700 video to the multi-action ground-truth labels for the human boxes within that video, treating these appended labels as pseudo-labels. We refer to this method as NWS.

\noindent
\textbf{Assignment-based Weak Supervision (AWS):} The NWS scheme outlined above presents two main issues: (1) not all actors in a Kinetics-700 video may be performing the global action assigned to that video (as shown in Figure \ref{fig:contrib}), and (2) the global action may not occur in the frames surrounding the AVA-annotated keyframe. To address these limitations, we introduce an enhanced approach based on self-training, named AWS. AWS training proceeds in two stages:
\begin{enumerate}
    \item \textbf{\textit{Initial Training with NWS:}} In the first step, we train the model using the NWS strategy.
    \item \textbf{\textit{Assigning Pseudolabels to Ground Truth Boxes:}} The NWS model is used to assign the global Kinetics-700 action class to the most relevant ground-truth human boxes in each Kinetics-700 video within AVA-Kinetics. Subsequently, the model is trained on these refined pseudolabels.
\end{enumerate}

Unlike traditional self-training approaches in object detection, we do not use the output boxes of the model as pseudoboxes for self-training. Instead, we use Hungarian matching on the output to allocate the global Kinetics-700 action class to the ideal ground-truth human box in each instance. This process ensures more accurate pseudolabels as shown in Figure \ref{fig:contrib} and improves the overall performance of action detection, as well as eliminating the additional uncertainty incurred by using pseudoboxes.

\subsection{Training Objective}
Our model is trained using a bipartite matching loss following transformer-based detection models \cite{detr, yolos, bmvit, star, tuber, vidt, stmixer}. For all predicted tripets $(\textbf{b}, \textbf{p}_{act}, \textbf{e}_{v})$, only actor scores $\textbf{p}_{act}$ and bounding box coordinates \textbf{b} are used for the initial Hungarian matching step to match predictions with ground truth labels, as the only object of interest is the human figure. Finally, our bipartite loss function is as follows:
$\mathcal{L}_{loss} = \lambda_{actor} {CE}_{actor} +
                      \lambda_{box} \mathcal{L}_{box} +
                      \lambda_{action} {CE}_{action}$                    
\noindent where ${CE}_{actor}$, $\mathcal{L}_{box}$ and ${CE}_{action}$ represent the actor classification loss, bounding box loss, and action classification loss, respectively.
Following OWL-ViT \cite{owlvit, owlv2}, each $\lambda$ is set to 2. More details on training can be found in the supplementary.

%% file: sec/4_exp.tex
\section{Experimental Setup}
\label{sec:exp_details}

\noindent \textbf{Implementation Details:}  
We initialize the video and text encoder from ViCLIP-B16 pretrained on InternVid-10m-FLT \cite{internvid}. More details on training configurations, video sampling strategy and hyperparameters can be found in the supplementary.

\noindent \textbf{GPT4-assisted Text Augmentation:} Following recent works \cite{llmsmeetvlms, openmixer} in text augmentation for open-vocabulary detection, we use GPT4 to generate 16 descriptors for each action class to alleviate generalization issues. During training, we randomly sample one descriptor for each action class that appears in a training batch. During evaluation, for a given action, we average the cosine similarity for all 16 descriptors to obtain one final cosine similarity for that action.

\noindent \textbf{Evaluation Metric:} We quantify our results using frame-level mean average precision (f-mAP) with Intersection-over-Union (IoU) threshold at 0.5 following previous works in action detection \cite{evad, bmvit, iclip}.

\subsection{Datasets}
We use six closed-set action detection datasets for our experiments: AVA \cite{ava} , AVA-Kinetics (AVA-K) \cite{avak},  UCF-101-24 \cite{ucf101}, JHMDB51-21 \cite{jhmdb}, Multisports \cite{multisports}, and UCF-MAMA \cite{ucfmama}.
\textbf{\textit{AVA}} contains 299 videos each lasting 15 minutes with keyframe annotations at every second. There are 80 atomic action classes in AVA, and the annotations are multi-label in nature.
\textbf{\textit{AVA-K}} contains 238,476 Kinetics-700 videos in addition to the original 299 AVA videos, annotated with 80 AVA classes in a similar manner. Similar to AVA, the annotations are multi-label in nature; currently, AVA and AVA-Kinetics are the only action detection datasets with multi-label human boxes. For both datasets, we only use AVA2.2 anntotations for AVA videos.

\textbf{\textit{UCF-101-24}} contains 2284 temporally untrimmed videos for training and 923 for testing distributed amongst 24 action classes. Annotations are in the form of spatiotemporal tubes.

\textbf{\textit{JHMDB51-21}} consists of 21 action classes split across 600 temporally trimmed videos for training and 300 for testing, with annotations in the form of spatiotemporal tubes.

\textbf{\textit{MultiSports}} consists of 4 sports and each sport consists of a set of fine-grained sport-specific actions, totaling to 66 action classes. It consists of 1574 untrimmed videos for training and 555 for validation. Similar to UCF-101-24 and JHMDB51-21, annotations in the form of spatiotemporal tubes.

\textbf{\textit{UCF-MAMA}} consists of high-resolution, temporally cropped videos from VIRAT \cite{virat} and MEVA \cite{meva} in a surveillance-style footage that depict humans and vehicles at a long range, totalling to 35 action classes. The annotations also include non-human actions, which we remove during training. Specifically, we remove vehicle actions (e.g. `\textit{Vehicle Turning Left}', `\textit{Vehicle Turning Right}'), reducing the number of action classes to 27.

\subsection{Training and Evaluation}
We evaluate our model using two settings: 1) \textit{Base-to-Novel:} A single dataset is partitioned into base and novel categories, following the approach established in previous open-vocabulary object detection works, such as \cite{ovr-cnn, ovod}. 2) \textit{Cross-dataset:} Training is performed on one dataset, while evaluation is conducted on separate downstream datasets. This setup aligns with the cross-dataset approach used in open-vocabulary object detection \cite{ovod}. We discuss the two setups for different datasets in detail below.

\noindent
\textbf{Base-to-Novel:} Following iCLIP \cite{iclip} and OpenMixer \cite{openmixer}, we use UCF-101-24 \cite{ucf101} or JHMDB \cite{jhmdb} and randomly split their videos into base classes for training and novel classes for zero-shot inference. The base-to-novel ratio is either 75\%-25\% or 50\%-50\%.  We use our weights pretrained on AVA-Kinetics + AWS before training our model in this setup.

\noindent
\textbf{The issue with base-to-novel:} JHMDB and UCF-101-24 are small action detection datasets with only 21 and 24 actions respectively. Splitting a set of novel actions from these datasets will result in fewer action classes for training and even fewer for evaluation, rendering it ill-posed. Instead of relying on this benchmark to evaluate open-vocabulary capabilities, we devise two cross-dataset schemes to evaluate on all action classes of any given downstream dataset without ever using them for training, similar to how all actions are evaluated in a closed-set setting.

\noindent
\textbf{Cross-Dataset:}
\textit{\textbf{1) AVA-Kinetics:}} Our primary contribution lies in this setting. Our model is trained on the 80 action classes from AVA, and we further employ NWS/AWS methods to increase the number of base action classes for training from 80 to over 700. For downstream evaluation, we evaluate on all action classes from UCF-101-24, JHMDB, MultiSports, and UCF-MAMA.
\textit{\textbf{2) UCF:JHMDB:}} We use the UCF-101-24 \cite{ucf101} classes as base classes for training and treat JHMDB \cite{jhmdb} classes as novel classes for zero-shot inference. We use this setup in our ablations.

\section{Results and Analysis}

\begin{table}[t]
\centering
\caption{\textbf{\textit{Open-vocabulary results on training with AVA-Kinetics}} without Kinetics-700 labels, with NWS and with AWS. * denotes that we only use human boxes/actions; non-human actors such as vehicles and their actions are removed.}
\renewcommand{\arraystretch}{1.06}
\scalebox{0.85}{
    \begin{tabular}{r cccc}
        \rowcolor{mygray} 
        \specialrule{1.5pt}{0pt}{0pt}
        Baseline & \multicolumn{1}{c}{UCF} & \multicolumn{1}{c}{JHMDB} & \multicolumn{1}{c}{MultiSports} & \multicolumn{1}{c}{UCF-MAMA*} \\
        \rowcolor{mygray} & f@0.5 & f@0.5 & f@0.5 & f@0.5 \\ 
        \hline
        LanguageBind \scriptsize{\cite{languagebind}} & 25.2 & 36.2 & 0.0 & 0.0 \\
        X-CLIP \scriptsize{\cite{xclip}} & 19.6 & 33.2 & 0.4 & 0.1 \\
        ViCLIP \scriptsize{\cite{internvid}} & 25.5 & 39.9 & 0.1 & 0.2 \\
        \hline
        \rowcolor{mygray} 
        \specialrule{1.5pt}{0pt}{0pt}
        Ours & \multicolumn{1}{c}{UCF} & \multicolumn{1}{c}{JHMDB} & \multicolumn{1}{c}{MultiSports} & \multicolumn{1}{c}{UCF-MAMA*} \\
        \rowcolor{mygray} & f@0.5 & f@0.5 & f@0.5 & f@0.5 \\
        SiA-B16 & 33.7 & 39.9 & \textbf{1.3} & 0.5 \\
        + NWS & 36.7 & 51.5 & 0.2 & 0.6 \\
        + AWS & \textbf{42.6} & \textbf{57.1} & 0.8 & \textbf{0.6} \\
        \specialrule{1.5pt}{0pt}{0pt}
    \end{tabular}}
\label{tab:inter_avak}
\vspace{-10pt}
\end{table}

\begin{table}[t]
    \centering
    \caption{\textbf{\textit{JHMDB base-to-novel results}} for both 75-25 and 50-50 splits (\%).}
    \renewcommand{\arraystretch}{1.06}
    \scalebox{0.9}{
        \begin{tabular}{r cc cccc cccc}
        \rowcolor{mygray} 
        \specialrule{1.5pt}{0pt}{0pt}
        Model & \multicolumn{2}{c}{75-25 Split} & \multicolumn{2}{c}{50-50 Split} \\
        \rowcolor{mygray} & Base@0.5 & Novel@0.5 & Base@0.5 & Novel@0.5 \\ 
        \hline\hline
        iCLIP \scriptsize{\cite{iclip}} & - & 66.8 & - & 45.2 \\
        OpenMixer \scriptsize{\cite{openmixer}} & - & 77.1 & - & - \\
        \hline
        SiA-B16 & 81.4 & \textbf{83.2} & 87.5 & \textbf{61.0} \\
        \specialrule{1.5pt}{0pt}{0pt}
        \end{tabular}}
\label{tab:intra_jhmdb}
\vspace{-10pt}
\end{table}

\begin{table}[t]
    \centering
    \caption{\textbf{\textit{UCF-101-24 base-to-novel results}} for both 75-25 and 50-50 splits (\%).}
    \renewcommand{\arraystretch}{1.06}
    \scalebox{0.9}{
        \begin{tabular}{r cc cccc cccc}
        \rowcolor{mygray} 
        \specialrule{1.5pt}{0pt}{0pt}
        Model & \multicolumn{2}{c}{75-25 Split} & \multicolumn{2}{c}{50-50 Split} \\
        \rowcolor{mygray} & Base@0.5 & Novel@0.5 & Base@0.5 & Novel@0.5 \\ 
        \hline\hline
        iCLIP \scriptsize{\cite{iclip}} & - & 72.5 & - & 60.3 \\
        \hline
        SiA-B16 & 97.0 & \textbf{97.1} & 94.7 & \textbf{75.1} \\ 
        \specialrule{1.5pt}{0pt}{0pt}
        \end{tabular}}
\label{tab:intra_ucf101}
\vspace{-10pt}
\end{table}

\begin{figure*} 
    \centering
    \includegraphics[scale=0.34]{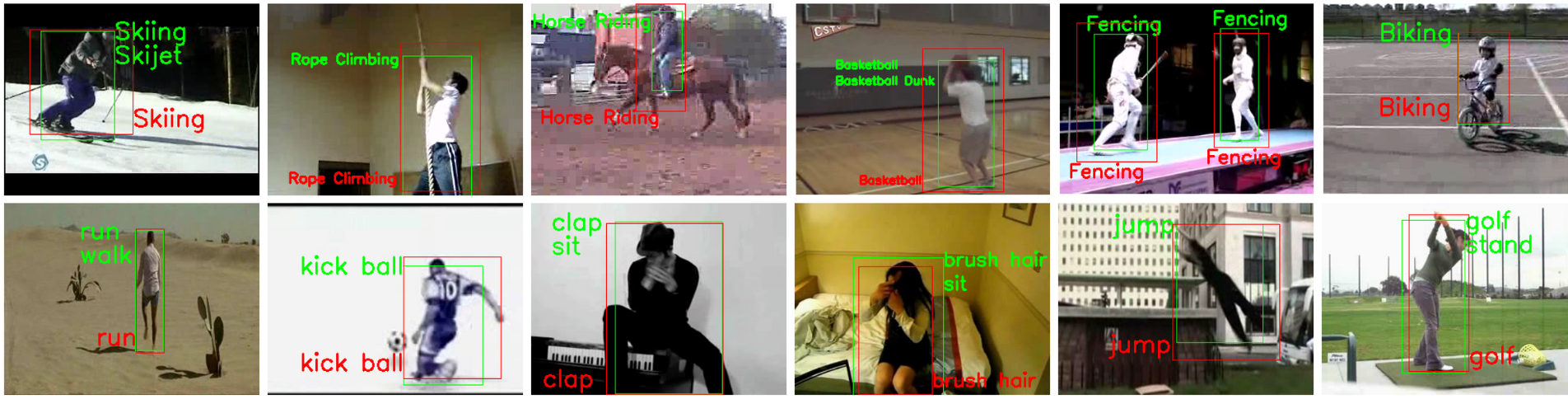}
    \caption{\textbf{\textit{Open-Vocabulary Qualitative Results}} on UCF-101-24 (1st row) and JHMDB (2nd row) from the output of our model, trained on AVA-Kinetics with AWS; this model is not trained on UCF-101-24 or JHMDB. Green boxes/labels denote predictions and red boxes/labels denote the ground truth. The multi-label nature of our model is also able to determine actions that are not labeled in the ground truth of JHMDB. Additionally, class confusion occurs for specific UCF-101-24 actions, notably `Basketball' and `Basketball Dunk'.
    }
    \label{fig:viz}
    \vspace{-10pt}
\end{figure*}

\textbf{Baseline:} In the absence of prior open-vocabulary results for our main benchmark in Table \ref{tab:inter_avak}, we devise a 3-stage baseline using off-the-shelf components. Our baseline is as follows: 1) For a given input clip, human detections are obtained for all frames using a pretrained human detector. 2) Human tubelets are built across the clip using ByteTrack \cite{bytetrack}. 3) The action associated with each human tubelet is obtained by cropping the tubelet from the input clip and classified by an off-the-shelf video-language model \cite{languagebind, xclip, internvid}.

\subsection{Open-Vocabulary Evaluation}

As shown in our new open-vocabulary benchmark in Table \ref{tab:inter_avak}, our model consistently exceeds the performance of the training-free baselines in all four downstream datasets, both with and without NWS and AWS.

For the base-to-novel benchmarks on JHMDB and UCF-101-24 in Table \ref{tab:intra_jhmdb} and \ref{tab:intra_ucf101} respectively, our method outperforms iCLIP on both 75:25 and 50:50 splits, as well as OpenMixer on the 75:25 split for JHMDB.

\subsection{Closed-Set Evaluation}
\label{sec:closedsetresults}

\begin{table}[t]
\centering
    \caption{\textbf{\textit{Closed-set results}} after full-finetuning on the downstream datasets. \textbf{*} denotes that we only use human actions.}
    \renewcommand{\arraystretch}{1.06}
    \scalebox{0.9}{
        \begin{tabular}{r c c c c}
        \rowcolor{mygray} 
        \specialrule{1.5pt}{0pt}{0pt}
        Model & UCF & JHMDB & MultiSports & UCF-MAMA \\
        \rowcolor{mygray} & f@0.5 & f@0.5 & f@0.5 & f@0.5 \\ 
        \hline\hline
        YOWO \cite{yowo} & 75.7 & 80.4 & - & - \\
        TubeR \cite{tuber} & 81.3 & - & - & - \\ 
        STMixer \cite{stmixer} & 83.7 & 86.7 & - & - \\
        YOWOv2 \cite{yowov2} & 87.0 & - & - & - \\
        HIT \cite{hit} & 84.8 & 83.8 & 33.3 & - \\
        EVAD \cite{evad} & 85.1 & 90.2 & - & - \\
        BMViT \cite{bmvit} & \textbf{90.7} & 88.4 & - & - \\
        STAR \cite{star} & 90.3 & \textbf{92.1} & \textbf{59.3} & - \\
        VCN-MA \cite{ucfmama} & - & - & - & 0.4 \\
        \hline
        SiA-B16 & 88.5 & 88.5 & 28.8 & \textbf{4.0*} \\
        \specialrule{1.5pt}{0pt}{0pt}
        \end{tabular}}
    \label{tab:closed-set}
\vspace{-10pt}
\end{table}

We show that our model also performs sufficiently in a closed-set setting by comparing against the latest closed-set action detection models. We initialize our model from AWS-pretrained weights and finetune them on each downstream closed-set action detection dataset. No text augmentation is applied and all available actions are passed into the text encoder to emulate closed-set training. Our results are shown in Table \ref{tab:closed-set}. Closed-set performance is higher than open-vocabulary, consistent with findings in image object detection \cite{vild, vlplm}.

\begin{figure*}
    \centering
    \includegraphics[scale=0.3]{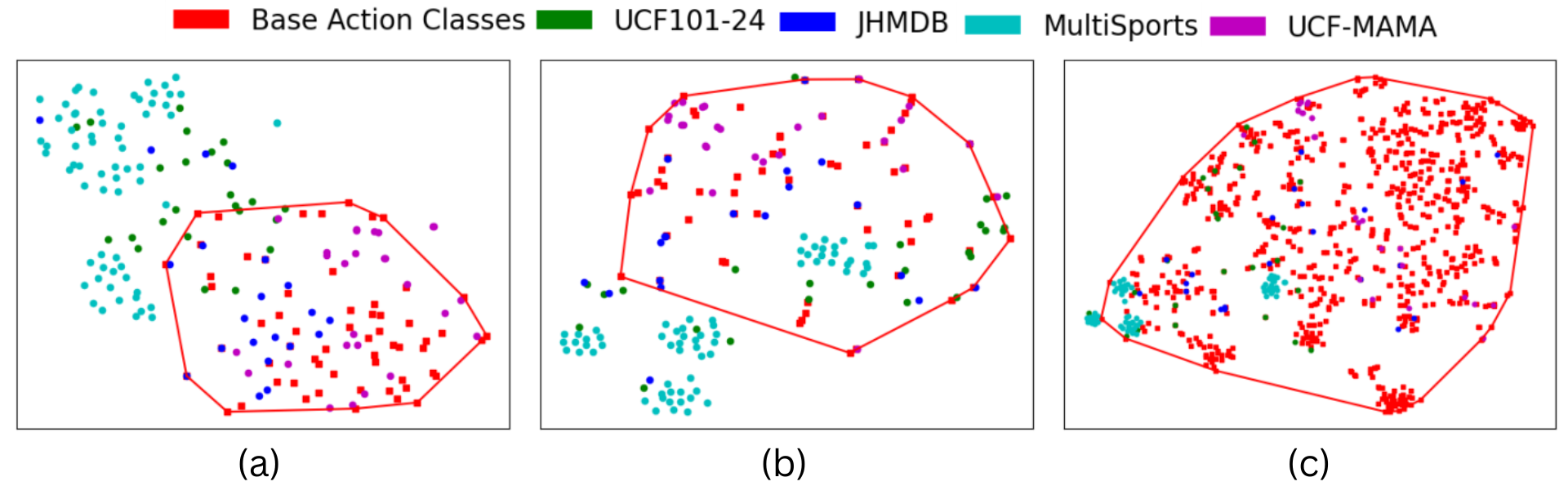}
    \caption{\textbf{\textit{t-SNE plot of text embeddings}} for our model trained on AVA-Kinetics with (a) frozen text encoder, (b) LoRA-finetuned text encoder, and (c) LoRA-finetuned text encoder with AWS to include the 700 actions from Kinetics-700. More downstream classes lie within the cluster of base (AVA) classes after LoRA-finetuning the text encoder, and even more lie within the AVA and Kinetics-700 classes after introducing AWS.
    }
    \label{fig:embeds}
\vspace{-10pt}
\end{figure*}

\begin{figure*}
    \centering
    \includegraphics[scale=0.34]{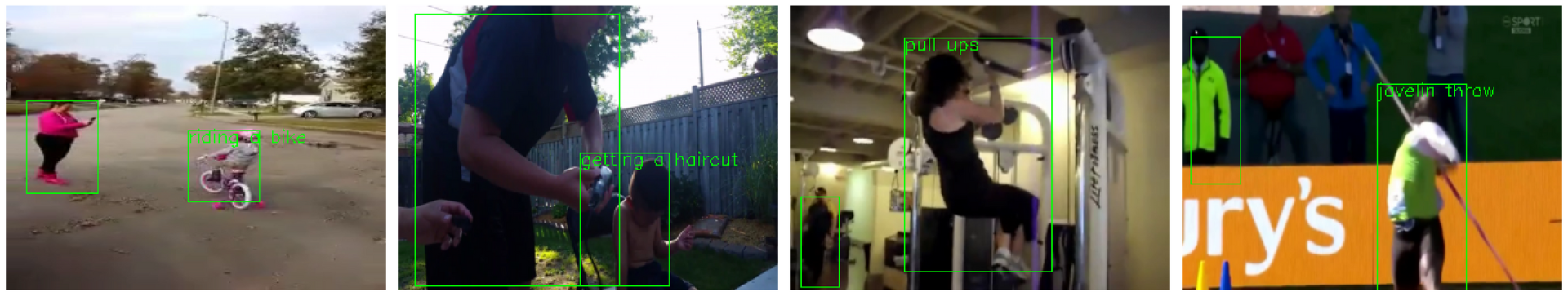}
    \caption{\textbf{\textit{Visualizations of AWS pseudolabel assignment}} showing multi-human instances where the global action has been allocated to the correct person.
    }
    \label{fig:awspredviz}
\end{figure*}

\subsection{Analysis of NWS and AWS}
\label{sec:wkanalysis}

\textbf{Quantitative comparison:} NWS results in a notable performance boost on UCF-101-24, JHMDB, and UCF-MAMA, as opposed to only training on 80 AVA actions.

While AWS cannot correct every single noisy pseudolabel from NWS, the performance increase compared to NWS suggests that most of the erroneous pseudolabels from NWS have been rectified, as shown in Table \ref{tab:inter_avak}.

\noindent
\textbf{Qualitative analysis:} We must highlight that AVA-Kinetics annotations for Kinetics-700 videos have 1.2 human boxes on average; most of the erroneous pseudolabels are caused by annotations with more than 1 person, which is only a small proportion of the dataset (28\%). Nevertheless, for certain videos, AWS is able to correct NWS pseudolabels as shown in Figure \ref{fig:awspredviz}.

\noindent
\textbf{Visualization of actions in embedding space:} From the t-SNE plots in Figure \ref{fig:embeds}, we can observe that introducing NWS and AWS to include Kinetics-700 actions alongside AVA actions further expands the convex hull of base action embeddings, and majority of the downstream actions from the aforementioned datasets lie within this hull.

\subsection{Ablations}

\begin{figure} 
    \centering
    \includegraphics[scale=0.45]{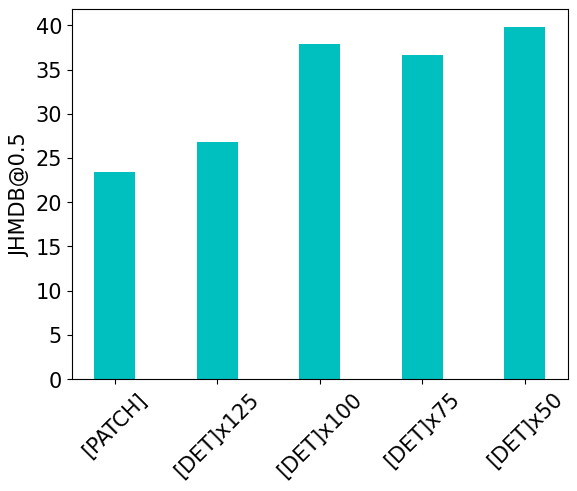}
    \caption{\textbf{\textit{Ablation:}} [PATCH] vs [DET] token regression on training with UCF-101-24 and evaluation on JHMDB.
    }
    \label{fig:bar_patchvsdet}
\end{figure}

\noindent \textbf{Encoder-only design: [PATCH] vs [DET] token regression:} As shown in Figure \ref{fig:bar_patchvsdet}, regressing [DET] tokens yields better downstream results than using [PATCH] tokens.

\begin{figure} 
    \centering
    \includegraphics[scale=0.45]{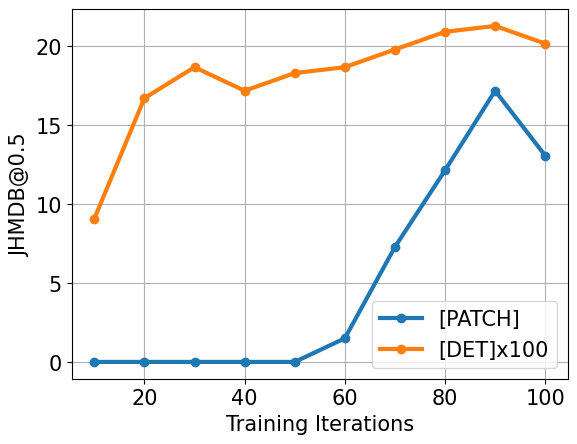}
    \caption{\textbf{\textit{Ablation:}} f-mAP@0.5 for JHMDB during the first 100 iterations of training on UCF-101-24.
    }
    \label{fig:patchvsdet}
\end{figure}

Furthermore, within the first 100 training iterations, the model with the [DET] token design demonstrates faster convergence compared to the model using only [PATCH] tokens, as shown in Figure \ref{fig:patchvsdet}.

In summary, the [DET] token scheme is a more effective approach for converting the video encoder of ViCLIP for action detection, and we finalize our design on this scheme.

\noindent \textbf{Impact of the number of [DET] tokens:} From Figure \ref{fig:patchvsdet}, we observe that increasing the number of [DET] tokens is detrimental to downstream performance. Nevertheless, to accommodate real-world use-cases where many people can be present in a surveillance-style footage such as UCF-MAMA, we choose to use 100 [DET] tokens as the default.

\noindent \textbf{Importance of finetuning the text encoder:} From Table \ref{tab:abla_txtlora} we find that introducing LoRA to the frozen text encoder yields a significant increase in performance, as opposed to keeping it frozen. This highlights the need to adapt the embeddings from the video-language pretrained text-encoder to be region specific.

From the t-SNE plots in Figure \ref{fig:embeds}, we observe that LoRA-finetuning the text encoder expands the convex hull of the base (AVA) action class embeddings, which starts to include more downstream actions from UCF-101-24, JHMDB, MultiSports and UCF-MAMA within the cluster of base action embeddings.

\noindent \textbf{GPT4-assisted text augmentation:} In Table \ref{tab:abla_gpt4}, we observe that the use of GPT4-generated descriptors for action classes yields a significant increase in detection performance compared to simply using class names.

\begin{table}[t]
    \centering
    \caption{\textbf{\textit{Ablation:}} Impact of LoRA-finetuning the text encoder.}
    \renewcommand{\arraystretch}{1.06}
    \scalebox{0.85}{
    \begin{tabular}{r ccccc}
        \rowcolor{mygray} 
        \specialrule{1.5pt}{0pt}{0pt}
        & & \multicolumn{1}{c}{Base} & \multicolumn{1}{c}{UCF-101-24} & \multicolumn{1}{c}{JHMDB} \\
        \rowcolor{mygray} Dataset & Text & f@0.5 & f@0.5 & f@0.5   \\ 
        \hline\hline
        AVA & Frozen & 10.2 & 5.1 & 6.1 \\
        AVA & LoRA & 20.1 & 19.0 & 34.4 \\
        \hline
        AVA-K & Frozen & 12.3 & 6.0 & 14.3 \\
        AVA-K & LoRA & 23.2 & 33.7 & 39.9 \\
        \specialrule{1.5pt}{0pt}{0pt}
        \end{tabular}}
\label{tab:abla_txtlora}
\vspace{-10pt}
\end{table}

\begin{table}[t]
    \centering
    \caption{\textbf{\textit{Ablation:}} Impact of GPT4-assisted text augmentation (with a frozen text encoder).}
    \renewcommand{\arraystretch}{1.06}
    \scalebox{0.85}{
    \begin{tabular}{ccccc}
        \rowcolor{mygray} 
        \specialrule{1.5pt}{0pt}{0pt}
        & & \multicolumn{1}{c}{Base} & \multicolumn{1}{c}{UCF-101-24} & \multicolumn{1}{c}{JHMDB} \\
        \rowcolor{mygray} Dataset & GPT4 & f@0.5 & f@0.5 & f@0.5  \\ 
        \hline\hline
        AVA & $\times$ & 7.4 & 0.6 & 2.7 \\
        AVA & \checkmark & 10.2 & 5.1 & 6.1 \\
        \hline
        AVA-K & $\times$ & 10.8 & 4.8 & 6.1  \\
        AVA-K & \checkmark & 12.3 & 6.0 & 14.3  \\
        \specialrule{1.5pt}{0pt}{0pt}
    \end{tabular}}
\label{tab:abla_gpt4}
\vspace{-10pt}
\end{table}

\section{Discussion}

\noindent \textbf{Weak-supervision: scaling the number of videos vs number of actions:} Unlike existing weakly supervised scaling methods in object detection for images which prioritize increasing the number of images, our approach focuses on expanding the number of action classes, as humans are the sole object of interest, which we achieve by adding more action labels to the already-annotated human boxes in an existing action detection dataset, providing our model with a more comprehensive understanding of potential actions without the need to increase the number of videos.

\noindent \textbf{Domain-specific actions:} Fine-grained actions from the MultiSports dataset require sport-specific domain knowledge (e.g., aerobic kick jump vs. aerobic straddle jump). Closed-set performance of our model on this dataset is significantly higher than open-vocabulary, as shown in Tables \ref{tab:inter_avak} and \ref{tab:closed-set}. We conclude that domain-specific actions are better handled with a closed-set training approach.

\noindent \textbf{The issue with single-action datasets:} Our model effectively detects multiple simultaneous actions, even in datasets annotated with only a single action label per instance (e.g., JHMDB), as shown in Figure \ref{fig:viz}. This highlights a fundamental flaw in such datasets: they impose an artificial constraint by labeling each video with only one action, despite real-world scenarios where multiple actions co-occur. This forces models to ignore secondary actions during evaluation. Additionally, single-action datasets that contain ambiguous label hierarchies (e.g., UCF-101-24 includes both "basketball" and "basketball dunk,") unfairly penalizes models that recognize broader activities but fail to predict the most specific label. Such inconsistencies distort performance metrics and hinder the development of truly generalizable models.

%% file: sec/5_conclusion.tex
\section{Conclusion}

In this work, we addressed the challenges of open-vocabulary action detection by introducing SiA, a single-stage, encoder-only model trained end-to-end for action detection and a weakly supervised training strategy that enables SiA to see more than 700 action classes during training, a significant departure from prior work that involve complicated adaptations to pretrained vision-language models for detection that sees less than 18 actions during training. Finally, we introduce a new cross-dataset benchmark to evaluate open-vocabulary action detection to replace the ill-posed base-to-novel benchmark on small action detection datasets used by prior works in zero-shot/open-vocabulary action detection, showing novel results to serve as baselines for future work.

\section{Acknowledgment}

This material is based upon work supported by the National Science Foundation (NSF) Accelerating Research Translation (ART) program under Grant No. 2331319. In addition, this research has benefited from the Microsoft Accelerating Foundation Models Research (AFMR) grant program.

Any opinions, findings, and conclusions or recommendations expressed in this material are those of the author(s) and do not necessarily reflect the views of the NSF or Microsoft.